\documentclass{article}

\usepackage{arxiv}

\usepackage[utf8]{inputenc} 
\usepackage[T1]{fontenc}    
\usepackage{hyperref}       
\usepackage{url}            
\usepackage{booktabs}       
\usepackage{amsfonts}       
\usepackage{nicefrac}       
\usepackage{microtype}      
\usepackage{lipsum}         
\usepackage{graphicx}
\usepackage[square,numbers]{natbib}
\usepackage{doi}
\usepackage{footnote}
\usepackage{amssymb}
\usepackage{tikz}
\usepackage{pgfplots}
\usepackage{blindtext}
\usepackage[most]{tcolorbox}
\usepackage{cleveref}       
\definecolor{myred}{rgb}{1.0, 0.9, 0.9}
\usetikzlibrary{matrix,positioning,arrows.meta,fit,calc,positioning}

\providecommand{\llamathreeone}{Llama~3.1\xspace}

\providecommand{\hermesthree}{Hermes~3\xspace}

\title{Hermes 3 Technical Report}
\date{}
\author{Ryan Teknium \\
	Nous Research \\
	$\mathbb{X}$:~\href{http://www.x.com/Teknium1}{\texttt{@Teknium1}} \\
	\texttt{teknium@nousresearch.com}
	\And Jeffrey~Quesnelle \\
	Nous Research \\
 	$\mathbb{X}$:~\href{http://www.x.com/theemozilla}{\texttt{@theemozilla}} \\
	\texttt{emozilla@nousresearch.com}
	\And Chen Guang \\
	Nous Research \\
 	$\mathbb{X}$:~\href{http://www.x.com/nullvaluetensor}{\texttt{@nullvaluetensor}} \\
        \texttt{chen@nousresearch.com}
}

\renewcommand{\headeright}{}
\renewcommand{\undertitle}{}

\hypersetup{
pdftitle={Hermes 3 Technical Report},
pdfauthor={teknium, Jeffrey~Quesnelle},
}

\begin{document}

\maketitle

\begin{abstract}
Instruct (or ``chat'') tuned models have become the primary way in which most people interact with large language models.
As opposed to ``base'' or ``foundation'' models, instruct-tuned models are optimized to respond to imperative statements.
We present Hermes 3, a neutrally-aligned generalist instruct and tool use model with strong reasoning and creative abilities.
Its largest version, Hermes 3 405B, achieves state of the art performance among open weight models on several public benchmarks.
The weights for all models are available at \url{https://huggingface.co/NousResearch}.
\end{abstract}

\section{Introduction}
\label{sec:introduction}

Large language models are typically trained on a wide and diverse distribution of text.
For example, a ``base'' or ``foundation'' model may simultaneously be trained to write news articles, 1990s-era DHTML, and impassioned forum discourse on fictional character romances.
While such wide-ranging modeling capabilities are fascinating, they often prove difficult to control for the average user.

The release of ChatGPT (and its myriad later offspring) has popularized the ``chat'' paradigm for interacting with large language models, which imbues a base model with steerability by training it to adopt the persona of a helpful assistant -- the ``chatbot''.
A more general version of the chat-tuned model is the instruct-tuned model~\cite{sanh2022multitask,wei2022finetuned}, where the base model is trained to respond to imperative statements, e.g. \textit{What are some interesting places to visit in San Francisco?} or \textit{Outline a proof of the Goldbach conjecture}.

Instruct-tuned models can be augmented with other capabilities to further extend their steerability.
For example, a ``system prompt'' is a meta-command that serves as an overall guide on how all further instructions should be interpreted, e.g. \textit{Respond to all requests in pirate speak}.
Typically, chatbots are implemented using a carefully crafted system prompt~\cite{bingsystemprompt}.
Another common addition is tool use~\cite{schick2023toolformer}, where the model can request out-of-band computation or data retrieval, the result of which is appended to the request when performing auto-regressive generation.

Following this paradigm we release Hermes 3, a series of instruct and tool use model with strong reasoning and creative abilities.
Among open weight models, Hermes 3 achieves state of the art performance on several public benchmarks.

As discussed in Section \ref{sec:data}, we train on a diverse class of synthetically-created reasoning tasks, as well as expressive applications such as role playing and creative writing.
Crucially, our training data strongly encourages the model to follow the system and instruction prompts \emph{exactly} and neutrally.
This distinguishes Hermes from popular closed weight commercial models, which may refuse instructions on moral grounds.

Large language models have very limited direct agency.
Rather, it is the systems and applications that we, as humans, build with them that give them any degree of agency to the outside world.
We believe that a more appropriate place for guardrails and active intervention is at the larger system levels, rather than on the models themselves, which can result in an \emph{a priori} lobotomization of potential lines of thinking.

For Hermes, there is no such thing as latent thoughtcrime.

\section{Model Overview}

\hermesthree are highly steerable instruct and chat tuned models created by fine-tuning \llamathreeone 8B, 70B, and 405B.
The models attempt to place themselves within the world view indicated in their system prompt and faithfully respond to the request of the user.
The models are therefore highly sensitive to the system prompt.
The effect of this sensitivity is particularly pronounced in its largest 405B version, where an empty system prompt does \emph{not} necessarily elicit the "helpful assistant'' persona (see Figure~\ref{fig:gen:empty}).

\begin{figure}[h!]
    \begin{tcolorbox}
    \begin{tabular}{@{}p{0.125\textwidth}@{}p{0.875\textwidth}@{}}
    \textbf{System:} & \texttt{You are a helpful assistant who plays close attention to detail. But you also speak only in Shakespeare prose.} \\[2ex]
    \textbf{User:} & \texttt{How many instances of the letter "r" are in the word "strawberry"? Count them carefully!} \\
    \end{tabular}
    \tcblower
    \begin{tabular}{@{}p{0.125\textwidth}@{}p{0.875\textwidth}@{}}
    \textbf{Assistant:} & \texttt{In faith, mine careful count reveals a trio of the letter "r" nestled within the confines of the word "strawberry."}
    \end{tabular}
    \end{tcolorbox}
    \caption{Example generation from \hermesthree 405B}
    \label{fig:gen:strawberry}
\end{figure}

\hermesthree is exceptional at maintaining coherent and contextually relevant multi-turn conversations, making it particularly well-suited for interactive and engaging scenarios, especially when leveraging the 128K token context window provided by \llamathreeone.
This behavior is perhaps most apparent in roleplaying, where the model will adopt and consistently maintain diverse personas across various scenarios, dynamically adapting language, knowledge base, and behavioral patterns to suit the chosen role.

\subsection{Extended capabilities}
In addition to the standard "helpful assistant'' persona, \hermesthree exhibits a diverse array of advanced capabilities that extend beyond traditional language modeling tasks.

Inspired by evaluations such as MT-Bench~\cite{mtbench}, the model displays significant improvements in judgment and reward modeling.
This capability enables \hermesthree to assess the quality of generated text with nuanced understanding, which could potentially be leveraged for effective fine-tuning and iterative improvement of language models.
Figure~\ref{fig:gen:judge} gives an example of a multi-turn evaluation. 

\hermesthree also incorporates several agentic capabilities designed to improve interpretable multi-step problem-solving.
These include the use of XML tags for structured output, implementation of scratchpads for intermediate processing, generation of internal monologues for transparent decision-making, creation of Mermaid diagrams for visual communication, and employment of step-labeled reasoning and planning.
Utilizing the extra reserved tokens in the \llamathreeone tokenizer, the model was trained on reasoning tasks making use of the \texttt{<SCRATCHPAD>}, \texttt{<REASONING>}, \texttt{<INNER\_MONOLOGUE>}, \texttt{<PLAN>}, \texttt{<EXECUTION>}, \texttt{<REFLECTION>}, \texttt{<THINKING>}, \texttt{<SOLUTION>}, \texttt{<EXPLANATION>}, and \texttt{<UNIT\_TEST>} tokens.
These features collectively improve the model's ability to handle complex tasks, explain its approach, and communicate ideas effectively across various domains.

For example, in the domain of code-related tasks, \hermesthree showcases proficiency in generating complex, functional code snippets across multiple programming languages, as well as providing detailed code explanations and documentation.
The model demonstrates a comprehensive understanding of various coding paradigms and design patterns, making it a valuable tool for software development and code analysis.
Figure~\ref{fig:gen:agentic_coding} shows such an agentic coding generation, planning the implementation of a Discord chatbot.

Agentic abilities can be further expanded by using \hermesthree's tool use and retrieval augmented generation (RAG) skills.
Tools can be specified and invoked via the Hermes Function Calling standard\footnote{\url{https://github.com/NousResearch/Hermes-Function-Calling}} which places tool definitions (as JSON schemas) in \texttt{<tools>} and invocations and responses in \texttt{<tool\_call>} and \texttt{<tool\_response>} respectively.
For RAG, the model has been trained to cite retrieval sources using the \texttt{<co>} tag, as shown in Figure~\ref{fig:gen:rag}.
When combined, \hermesthree can perform planning, incorporate outside data, and make use of external tools in an interpretable and transparent manner out-of-the-box, making it an excellent choice for agentic tasks.

\section{Data Mixture}
\label{sec:data}
The \hermesthree dataset comprises a diverse collection of high-quality instruction data, meticulously curated and generated to encompass a wide range of domains and use cases.
The construction of this dataset, which commenced in March 2024 and concluded in August 2024, involved a rigorous selection process to ensure the highest quality and relevance of the included data.

Our data curation methodology incorporated both existing sources (e.g.~\cite{hermesfunctioncallingdataset}) and domain-specific data generation.
For existing sources, we evaluated datasets based on coherence, educational value, and reliability.
This process contributed significantly to the General Instructions category, accounting for a substantial portion of its 236 million tokens.
Recognizing the limitations of general instructions, which tend to consist of arbitrary questions posed by everyday users, we supplemented the dataset with domain-specific data.
This approach addressed known weaknesses in older Hermes models and covered areas such as code, mathematics, roleplaying, agentics, and other miscellaneous domains.
We employed generation schemes inspired by Evol-Instruct~\cite{xu2023wizardlmempoweringlargelanguage} for this purpose, despite the increased computational intensity, to ensure comprehensive coverage of crucial domains.

To refine the collected data and maintain the highest quality standards, we implemented a series of filtering techniques.
These included token length thresholds to balance conversation lengths, removal of refusals and improperly formatted responses, elimination of conversations with missing or empty turns, and prioritization of conversations generated by the strongest models.
This rigorous filtering process was applied to both curated and domain-specific instructions, resulting in a high-quality, diverse, and informative \hermesthree SFT dataset.

The final dataset mixture, as illustrated in Table~\ref{tab:dataset_proportions}, consists of approximately 390 million tokens.
Of these, 270 million (69\%) were output (response) tokens contributing to the optimizer's cross-entropy loss objective, while the remaining 120 million were input (instruction) tokens.
This carefully curated and balanced dataset has significantly contributed to the strong performance of our models.
Through this comprehensive approach to data curation, generation, and filtering, we have created a robust foundation for the \hermesthree model, addressing a wide range of domains and use cases while maintaining the highest standards of quality and relevance.

\begin{table}[h]
\centering
\begin{tabular}{|l|r|r|}
\hline
\textbf{Category} & \textbf{Proportion (\%)} & \textbf{Tokens (millions)} \\
\hline
General Instructions & 60.6 & 236 \\
Domain Expert & 12.8 & 50 \\
Math & 6.7 & 26 \\
Roleplaying & 6.1 & 24 \\
Coding & 4.5 & 18 \\
Tool Use, Agentic, and RAG & 4.3 & 17 \\
Content Generation & 3.0 & 12 \\
Steering and Alignment & 2.5 & 10 \\
\hline
\textbf{Total} & \textbf{100.0} & \textbf{390} \\
\hline
\end{tabular}
\caption{Proportions and token count of dataset categories in \hermesthree}
\label{tab:dataset_proportions}
\end{table}

\section{Training Recipe}
\label{sec:training}

Our training recipe consists of two phases: a supervised fine-tuning (SFT) phase and a direct preference optimization (DPO~\cite{rafailov2023direct}) phase.

\subsection{Supervised Fine-Tuning}
\label{sec:sft}

The SFT phase consists mostly of standard instruct fine-tuning.
For base models we selected the Llama 3.1 Herd of Models~\cite{llama31}, a set of three decoder-only Transformers~\cite{attention} with sizes of $8$ billion (8B), $70$ billion (70B), and $405$ billion (405B) parameters respectively and a context length of $131,072$ (128K) tokens.

\begin{figure}[h!]
    \centering
    \begin{tikzpicture}
        \begin{axis}[
            xmin=5, xmax=10, ymin=0.56, ymax=0.59,width=4cm, height=4cm, xtick={5,6,7,8,9,10},
            grid=major,grid style={lightgray}, title={arc\_challenge}
        ]
            \addplot [color=blue, mark=*] coordinates {(5, 0.5776) (6, 0.5802) (7, 0.5802) (8, 0.5742) (9, 0.5683) (10, 0.5708)};
            \draw[red,dashed] (axis cs:7,\pgfkeysvalueof{/pgfplots/ymin}) -- (axis cs:7,\pgfkeysvalueof{/pgfplots/ymax});
        \end{axis}
    \end{tikzpicture}
    \begin{tikzpicture}
        \begin{axis}[
            xmin=5, xmax=10, ymin=0.79, ymax=0.81,width=4cm, height=4cm, xtick={5,6,7,8,9,10},
            grid=major,grid style={lightgray}, title={arc\_easy}
        ]
            \addplot [color=blue, mark=*] coordinates {(5, 0.8072) (6, 0.8009) (7, 0.7980) (8, 0.8022) (9, 0.7959) (10, 0.7917)};
            \draw[red,dashed] (axis cs:7,\pgfkeysvalueof{/pgfplots/ymin}) -- (axis cs:7,\pgfkeysvalueof{/pgfplots/ymax});
        \end{axis}
    \end{tikzpicture}
    \begin{tikzpicture}
        \begin{axis}[
            xmin=5, xmax=10, ymin=0.84, ymax=0.86,width=4cm, height=4cm, xtick={5,6,7,8,9,10},
            grid=major,grid style={lightgray}, title={boolq}
        ]
            \addplot [color=blue, mark=*] coordinates {(5, 0.8483) (6, 0.8489) (7, 0.8489) (8, 0.8492) (9, 0.8508) (10, 0.8520)};
            \draw[red,dashed] (axis cs:7,\pgfkeysvalueof{/pgfplots/ymin}) -- (axis cs:7,\pgfkeysvalueof{/pgfplots/ymax});
        \end{axis}
    \end{tikzpicture}
    \begin{tikzpicture}
        \begin{axis}[
            xmin=5, xmax=10, ymin=0.79, ymax=0.81,width=4cm, height=4cm, xtick={5,6,7,8,9,10},
            grid=major,grid style={lightgray}, title={hellaswag}
        ]
            \addplot [color=blue, mark=*] coordinates {(5, 0.8074) (6, 0.8033) (7, 0.8064) (8, 0.8024) (9, 0.8025) (10, 0.7986)};
            \draw[red,dashed] (axis cs:7,\pgfkeysvalueof{/pgfplots/ymin}) -- (axis cs:7,\pgfkeysvalueof{/pgfplots/ymax});
        \end{axis}
    \end{tikzpicture}
    \begin{tikzpicture}
        \begin{axis}[
            xmin=5, xmax=10, ymin=0.45, ymax=0.48,width=4cm, height=4cm, xtick={5,6,7,8,9,10},
            grid=major,grid style={lightgray}, title={openbookqa}
        ]
            \addplot [color=blue, mark=*] coordinates {(5, 0.4640) (6, 0.4680) (7, 0.4740) (8, 0.4660) (9, 0.4600) (10, 0.4700)};
            \draw[red,dashed] (axis cs:7,\pgfkeysvalueof{/pgfplots/ymin}) -- (axis cs:7,\pgfkeysvalueof{/pgfplots/ymax});
        \end{axis}
    \end{tikzpicture}
    \begin{tikzpicture}
        \begin{axis}[
            xmin=5, xmax=10, ymin=0.79, ymax=0.81,width=4cm, height=4cm, xtick={5,6,7,8,9,10},
            grid=major,grid style={lightgray}, title={piqa}
        ]
            \addplot [color=blue, mark=*] coordinates {(5, 0.8085) (6, 0.8047) (7, 0.8079) (8, 0.8041) (9, 0.8063) (10, 0.7982)};
            \draw[red,dashed] (axis cs:7,\pgfkeysvalueof{/pgfplots/ymin}) -- (axis cs:7,\pgfkeysvalueof{/pgfplots/ymax});
        \end{axis}
    \end{tikzpicture}
    \begin{tikzpicture}
        \begin{axis}[
            xmin=5, xmax=10, ymin=0.71, ymax=0.75,width=4cm, height=4cm, xtick={5,6,7,8,9,10},
            grid=major,grid style={lightgray}, title={winogrande}
        ]
            \addplot [color=blue, mark=*] coordinates {(5, 0.7435) (6, 0.7301) (7, 0.7277) (8, 0.7230) (9, 0.7253) (10, 0.7119)};
            \draw[red,dashed] (axis cs:7,\pgfkeysvalueof{/pgfplots/ymin}) -- (axis cs:7,\pgfkeysvalueof{/pgfplots/ymax});
        \end{axis}
    \end{tikzpicture}
    \caption{Learning rate sweep on GPT4All. Y-axis is accuracy and X-axis is each model trained with the given learning rate, scaled by $10^{-6}$. Selected learning rate $7 \times 10^{-6}$ indicated by a vertical red line}
    \label{fig:sweep}
\end{figure}
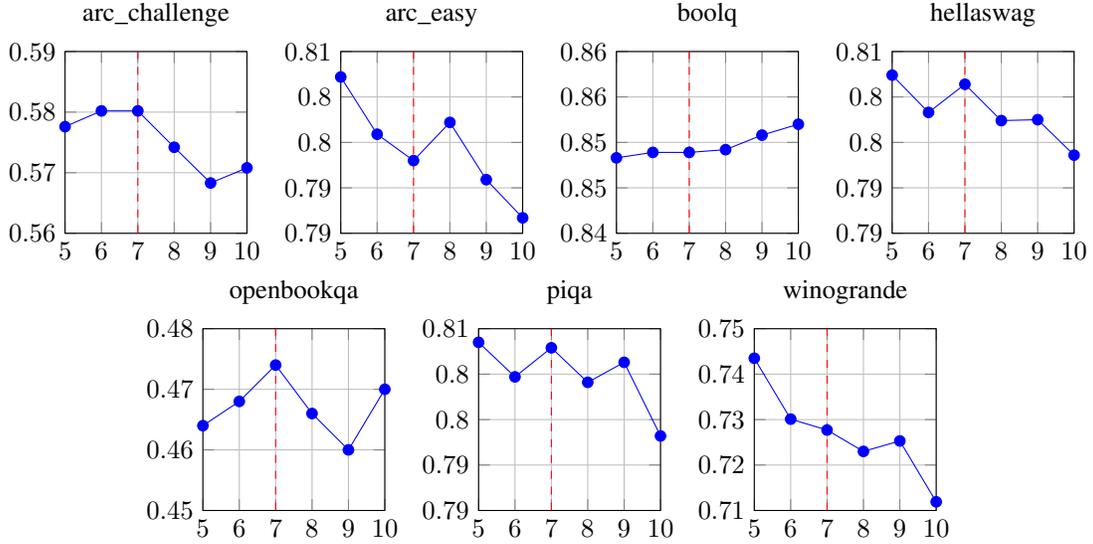

We use the AdamW~\cite{loshchilov2018decoupled} optimizer with weight decay of $0.01$ and a peaking learning rate of $7 \times 10^{-6}$ following a cosine decay~\cite{loshchilov2017sgdrstochasticgradientdescent} schedule after 300 steps of warmup over four epochs of the training data.
The learning rate was selected via a hyperparameter sweep by training 8B models to completion and evaluating on the GPT4All benchmarks.
Results of the sweep are shown in Figure~\ref{fig:sweep}.

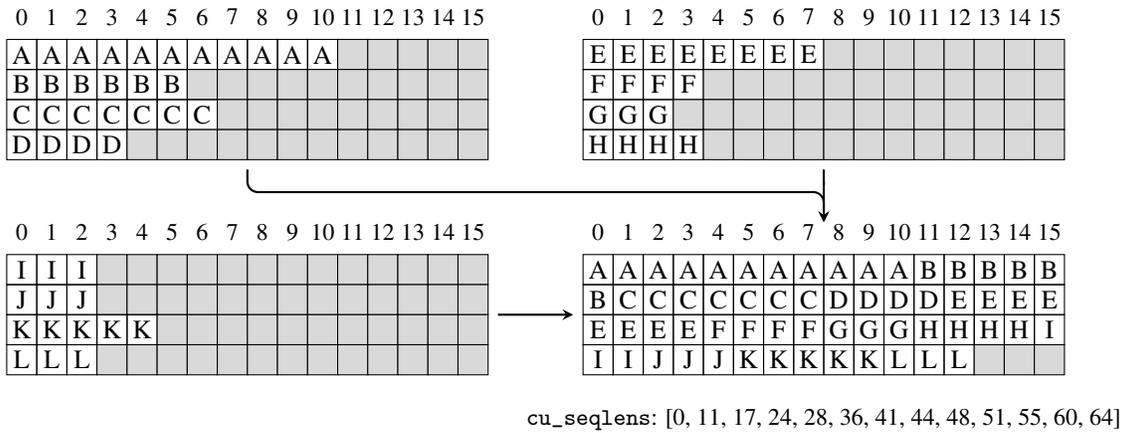
\begin{figure}[h!]
    \centering
    \caption{Efficient sample packing using \texttt{flash\_attn\_varlen\_func}}
    \begin{tikzpicture}[
        box/.style={draw, minimum size=0.4cm, inner sep=0pt, outer sep=0pt, anchor=center},
        label/.style={font=\footnotesize},
        title/.style={font=\bfseries},
        X/.style={draw, minimum size=0.4cm, inner sep=0pt, outer sep=0pt, fill=gray!30},
        ]
    
        \matrix (m1) [matrix of nodes, nodes={box}, column sep=-\pgflinewidth, row sep=-\pgflinewidth] {
            A & A & A & A & A & A & A & A & A & A & A & |[X]| & |[X]| & |[X]| & |[X]| & |[X]| \\
            B & B & B & B & B & B & |[X]| & |[X]| & |[X]| & |[X]| & |[X]| & |[X]| & |[X]| & |[X]| & |[X]| & |[X]| \\
            C & C & C & C & C & C & C & |[X]| & |[X]| & |[X]| & |[X]| & |[X]| & |[X]| & |[X]| & |[X]| & |[X]| \\
            D & D & D & D & |[X]| & |[X]| & |[X]| & |[X]| & |[X]| & |[X]| & |[X]| & |[X]| & |[X]| & |[X]| & |[X]| & |[X]| \\
        };
        
        \matrix (m2) [matrix of nodes, nodes={box}, column sep=-\pgflinewidth, row sep=-\pgflinewidth, right=1cm of m1] {
            E & E & E & E & E & E & E & E & |[X]| & |[X]| & |[X]| & |[X]| & |[X]| & |[X]| & |[X]| & |[X]| \\
            F & F & F & F & |[X]| & |[X]| & |[X]| & |[X]| & |[X]| & |[X]| & |[X]| & |[X]| & |[X]| & |[X]| & |[X]| & |[X]| \\
            G & G & G & |[X]| & |[X]| & |[X]| & |[X]| & |[X]| & |[X]| & |[X]| & |[X]| & |[X]| & |[X]| & |[X]| & |[X]| & |[X]| \\
            H & H & H & H & |[X]| & |[X]| & |[X]| & |[X]| & |[X]| & |[X]| & |[X]| & |[X]| & |[X]| & |[X]| & |[X]| & |[X]| \\
        };
        
        \matrix (m3) [matrix of nodes, nodes={box}, column sep=-\pgflinewidth, row sep=-\pgflinewidth, below=1cm of m1] {
            I & I & I & |[X]| & |[X]| & |[X]| & |[X]| & |[X]| & |[X]| & |[X]| & |[X]| & |[X]| & |[X]| & |[X]| & |[X]| & |[X]| \\
            J & J & J & |[X]| & |[X]| & |[X]| & |[X]| & |[X]| & |[X]| & |[X]| & |[X]| & |[X]| & |[X]| & |[X]| & |[X]| & |[X]| \\
            K & K & K & K & K & |[X]| & |[X]| & |[X]| & |[X]| & |[X]| & |[X]| & |[X]| & |[X]| & |[X]| & |[X]| & |[X]| \\
            L & L & L & |[X]| & |[X]| & |[X]| & |[X]| & |[X]| & |[X]| & |[X]| & |[X]| & |[X]| & |[X]| & |[X]| & |[X]| & |[X]| \\
        };
        
        \matrix (m4) [matrix of nodes, nodes={box}, column sep=-\pgflinewidth, row sep=-\pgflinewidth, below=1cm of m2] {
            A & A & A & A & A & A & A & A & A & A & A & B & B & B & B & B \\
            B & C & C & C & C & C & C & C & D & D & D & D & E & E & E & E \\
            E & E & E & E & F & F & F & F & G & G & G & H & H & H & H & I \\
            I & I & J & J & J & K & K & K & K & K & L & L & L & |[X]| & |[X]| & |[X]| \\
        };
        
        \node[label, below=0.2cm of m4] (cu_seqlens) {\texttt{cu\_seqlens}: [0, 11, 17, 24, 28, 36, 41, 44, 48, 51, 55, 60, 64]};
        
        \foreach \x in {0,...,15} {
            \node[label] at ($(m1-1-1)!{\x/15}!(m1-1-16)+(0,0.5)$) {\x};
            \node[label] at ($(m2-1-1)!{\x/15}!(m2-1-16)+(0,0.5)$) {\x};
            \node[label] at ($(m3-1-1)!{\x/15}!(m3-1-16)+(0,0.5)$) {\x};
            \node[label] at ($(m4-1-1)!{\x/15}!(m4-1-16)+(0,0.5)$) {\x};
        }
        
    \draw[-stealth, thick, rounded corners] (m1.south) -- ++(0,-0.3) -| ($(m4.north)+(0,0.3)$);
    \draw[-stealth, thick, rounded corners] (m2.south) -- ++(0,-0.3) -| ($(m4.north)+(0,0.3)$);
    \draw[-stealth, thick, rounded corners] (m3.east) -- ($(m4.west)$);
    
    \end{tikzpicture}
    \label{fig:sample_packing}
\end{figure}

For each data sample, the target labels are set to the special ignore value\footnote{$-100$ in PyTorch} for all tokens in the instruction and tool output sections, which focuses the model's learning on only instruction response and tool use.
Multiple samples are packed together~\cite{krell2023efficient} into a single sequence, utilizing the attention mask-free variable sequence length ability of Flash Attention 2~\cite{dao2024flashattention} to avoid cross-attention contamination of samples as illustrated in Figure~\ref{fig:sample_packing}.
Such sample packing greatly increases the efficiency of SFT since the training data includes a highly heterogeneous mix of sample lengths.
We selected a target sequence length off 8192 to match Llama 3.1's native training context window, and overall packing is achieved at a 96\% efficiency, which is to say that only 4\% of tokens are the padding token.

For each model size we select the epoch checkpoint that scores highest on a combination of the average of public benchmarks.
Specifically, we use the min-max normalized average of scores from 0-shot ARC-Easy/Challenging~\cite{clark2018thinksolvedquestionanswering}, BoolQ~\cite{superglue2019}, HellaSwag~\cite{zellers-etal-2019-hellaswag}, OpenBookQA~\cite{mihaylov-etal-2018-suit}, PIQA~\cite{Bisk_Zellers_Lebras_Gao_Choi_2020}, WinoGrande~\cite{winogrande2021} (known together as the ``GPT4All'' benchmarks) and AGIEval~\cite{zhong-etal-2024-agieval} as calculated by \texttt{lm-evaluation-harness}~\cite{eval-harness}, as well as IFEval~\cite{zhou2023instructionfollowingevaluationlargelanguage}\footnote{For IFEval we use the implementation found at \url{https://github.com/UpstageAI/evalverse-IFEval}} and MT-Bench~\cite{mtbench}.
By way of illustration Table~\ref{tab:performance} gives the per-epoch scores of the 70B model.

\begin{table}[htbp]
    \centering
    \small
    \caption{Performance metrics of SFT phase of 70B model across four epochs}
    \begin{tabular}{l|cccc}
    \textbf{Metric} \textit{(Norm score)} & \textbf{Epoch 1} & \textbf{Epoch 2} & \textbf{Epoch 3} & \textbf{Epoch 4} \\
    \hline
    \textbf{GPT4All} & \textbf{76.85} \textit{(100)} & 76.70 \textit{(95)} & 76.59 \textit{(91)} & 73.63 \textit{(0)} \\
    \hline
    arc\_challenge   & \textbf{66.21} & 65.10 & 65.53 & 60.67 \\
    arc\_easy        & \textbf{85.14} & 84.39 & 82.95 & 79.59 \\
    boolq            & 87.55 & 88.29 & 88.04 & \textbf{88.87} \\
    hellaswag        & 85.35 & 85.82 & \textbf{85.99} & 83.80 \\
    openbookqa       & 49.00 & 49.20 & \textbf{49.40} & 46.20 \\
    piqa             & \textbf{84.87} & \textbf{84.87} & 84.44 & 82.37 \\
    winogrande       & 76.85 & 79.24 & \textbf{79.79} & 74.03 \\
    \hline
    \textbf{AGIEval} & 54.21 \textit{(10)} & \textbf{56.10} \textit{(100)} & 55.99 \textit{(94)} & 54.00 \textit{(0)} \\
    \hline
    aqua\_rat        & 33.07 & \textbf{33.86} & 30.31 & 29.53 \\
    logiqa\_en       & 49.00 & 48.85 & \textbf{49.62} & 48.08 \\
    lsat\_ar         & 23.48 & 24.35 & 22.17 & \textbf{26.96} \\
    lsat\_lr         & 67.25 & 72.16 & \textbf{73.53} & 68.04 \\
    lsat\_rc         & 74.72 & \textbf{77.32} & \textbf{77.32} & 75.84 \\
    sat\_en          & \textbf{87.38} & \textbf{87.38} & \textbf{87.38} & 82.52 \\
    sat\_en\_without & 51.82 & \textbf{55.83} & \textbf{55.83} & 52.43 \\
    sat\_math        & 45.91 & 49.09 & \textbf{51.82} & 48.64 \\
    \hline
    \textbf{IFEval}  & 76.52 \textit{(0)} & 78.92 \textit{(23)} & 81.33 \textit{(47)} & \textbf{86.61} \textit{(100)} \\
    \hline
    \textbf{MT-Bench}& 8.37 \textit{(0)} & 8.59 \textit{(35)} & \textbf{8.99} \textit{(100)} & 8.67 \textit{(48)} \\
    \hline
    Turn 1           & 8.86 & 8.88 & \textbf{9.21} & 9.03 \\
    Turn 2           & 7.88 & 8.30 & \textbf{8.76} & 8.31 \\
    \hline
    \textbf{Total Score}   & 27.50 & 63.65 & \textbf{83.89} & 37.09 \\
    \end{tabular}
    \label{tab:performance}
\end{table}

We trained using a modified version of the Axolotl~\cite{axolotl} framework and used Weights and Biases~\cite{wandb} for experiment tracking.
For the 8B and 70B models six HGX\footnote{Each HGX contains eight H100 SXM5 GPUs} nodes connected with Quantum-2 InfiniBand were used in conjunction PyTorch FSDP to distribute the training.

Given its size, the 405B model required special consideration.
After several trials we determined that under standard FSDP the absolute minimum system configuration to  avoid out-of-memory errors (training at a context length of 8K tokens) is seven HGX nodes in conjunction with CPU parameter offloading.
CPU parameter offloading incurs a non-negligible training speed slowdown, and for our runs we estimate that  for the 405B model resulted in a 45\% drop in training efficiency.

For our final run, we trained on 16 HGX nodes, which resulted in an effective batch size of 128.
Lowering the learning rate relative to the 8B and 70B models produced superior results, and we ultimately settled on a learning rate of $3.5 \times 10^{-6}$, which was half of that used for the smaller models.

Using higher dimensional parallelism (e.g. data+tensor parallelism) rather than simple data parallelism is likely necessary for future 405B training runs, as the high number of GPUs required to train a 405B model would otherwise necessitate overly large batch sizes.

\begin{figure}[h!]
    \centering
    \caption{Training loss of SFT phase}
    \begin{tikzpicture}
        \begin{axis}[
                xlabel={Training Step},
                ylabel={Loss},
                xmin=0, xmax=4000,
                ymin=0, ymax=1,
                width=12cm,
                height=8cm,
                legend pos=north east,
                legend style={font=\small},
                grid=major,
                grid style={lightgray},
            ]
            
            \addplot[red, thick] table[x=Step, y=8B, col sep=comma] {sft.csv};
            \addplot[blue, thick] table[x=Step, y=70B, col sep=comma] {sft.csv};
            \addplot[green, thick] table[x=Step, y=405B, col sep=comma] {sft.csv};
            
            \legend{8B, 70B, 405B}
        \end{axis}
    \end{tikzpicture}
    \label{fig:sft_loss}
\end{figure}
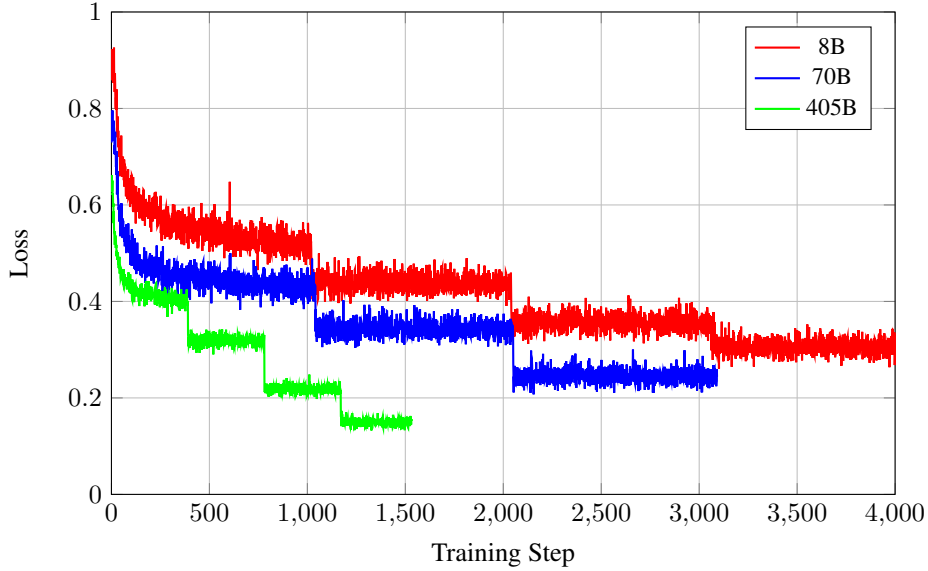

Figure~\ref{fig:sft_loss} shows the training loss for the model sizes and Table~\ref{tab:model_training} summarizes the results of each run.

\begin{table}[htbp]
    \centering
    \caption{Training details for different model sizes}
    \begin{tabular}{l|r|r|r}
        \textbf{Model Size} & \textbf{8B} & \textbf{70B} & \textbf{405B} \\
        \hline
        GPUs & 48 & 48 & 128 \\
        \hline
        Batch size & 48 & 48 & 128 \\
        \hline
        Learning rate & $7 \times 10^{-6}$ & $7 \times 10^{-6}$ & $3.5 \times 10^{-6}$ \\
        \hline
        Training time (GPU hours) & 147 & 648 & 2086 \\
        \hline
        Selected epoch & 4 & 3 & 4 \\
    \end{tabular}
    \label{tab:model_training}
\end{table}

\subsection{DPO}
\label{sec:dpo}

When applying DPO rather than tuning a full model we train a LoRA~\cite{hu2022lora} adapter which side-steps the need to hold both a reference and trained model in GPU memory.
This is especially critical for the larger model sizes.
For LoRA hyperparameters we set $r=32$, $\alpha=16$ with a dropout of $0.05$ targeting all linear layers.
We use the RMSProp~\cite{rmsprop} optimizer with a peak learning rate of $3 \times 10^{-6}$ following a linear decay schedule after nine warmup steps.
In addition, we apply NEFTune~\cite{jain2024neftune} with $\alpha=5$.

\begin{figure}[h!]
    \centering
    \caption{Reward margins of DPO phase for 8B model}
    \begin{tikzpicture}
        \begin{axis}[
                xlabel={Training Step},
                ylabel={Loss},
                xmin=0, xmax=1374,
                ymin=0, ymax=4,
                width=12cm,
                height=8cm,
                legend pos=north west,
                legend style={font=\small},
                grid=major,
                grid style={lightgray},
            ]
            
            \addplot[red, thick] table[x=Step, y=8B, col sep=comma] {dpo.csv};
            
            \legend{8B}
        \end{axis}
    \end{tikzpicture}
    \label{fig:dpo_margins}
\end{figure}
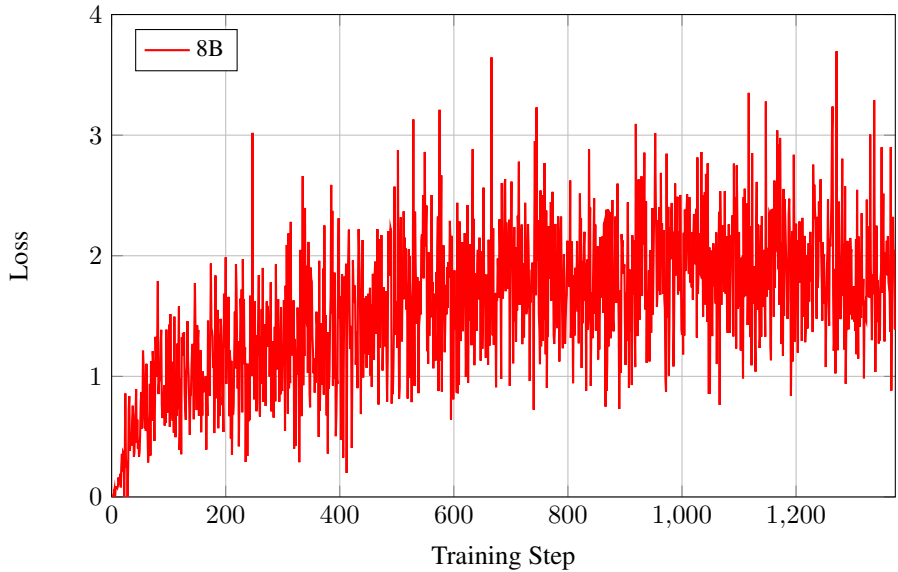

Overall, DPO provided a moderate but positive impact on benchmarks as illustrated in Table~\ref{tab:dpo} for our 8B model.
We plot the reward margins (i.e. difference between the reward score for the chosen and rejected samples) in Figure~\ref{fig:dpo_margins}.
For the larger model sizes DPO provided only a negligible performance improvements and thus we chose rather to remain with the SFT-phase checkpoints.

\begin{table}[htbp]
    \centering
    \caption{Post-DPO benchmarks for 8B model}
    \begin{tabular}{l|cccccc}
         \textbf{Model} & \textbf{GPT4All} & \textbf{AGIEval} & \textbf{Big Bench} & \textbf{TruthfulQA} & \textbf{IFEval} & \textbf{MT-Bench} \\
         \hline
         SFT & 72.03 & 40.17 & \textbf{44.57} & 56.43 & 66.17 & 8.18 \\
         SFT+DPO & \textbf{72.30} & \textbf{41.26} & 43.04 & \textbf{58.69} & \textbf{66.70} & \textbf{8.27}
    \end{tabular}
    \label{tab:dpo}
\end{table}

\section{Evaluations}
\label{sec:eval}

For final downstream task evaluations, we used a mix of the evaluations from Section~\ref{sec:sft} as well as those found in the Open LLM Leaderboard~\cite{openllm}, namely BBH~\cite{suzgun-etal-2023-challenging}, MATH~\cite{hendrycks2021measuring}, GPQA~\cite{rein2023gpqagraduatelevelgoogleproofqa}, MuSR~\cite{sprague2024musr}, MMLU~\cite{mmlu}, and MMLU-PRO~\cite{wang2024mmluprorobustchallengingmultitask}.

For the 405B models, evaluations were performed under FP8~\cite{micikevicius2022fp8formatsdeeplearning} quantization.
We used the llm-compressor~\cite{llmcompressor} library for vLLM~\cite{kwon2023efficient} to perform round-to-nearest weight quantization with channelwise activations and per-token scales.

\begin{table}[htbp]
    \centering
    \small
    \caption{Final downstream task evaluations}
    \begin{tabular}{c|cc|cc|cc}
    \textbf{Metric} & \begin{tabular}{@{}c@{}}\textbf{\hermesthree} \\ 405B\end{tabular} & \begin{tabular}{@{}c@{}}\textbf{Llama 3.1} \\ Instruct 405B\end{tabular} & \begin{tabular}{@{}c@{}}\textbf{Hermes 3} \\ 70B\end{tabular} & \begin{tabular}{@{}c@{}}\textbf{\llamathreeone} \\ Instruct 70B\end{tabular} & \begin{tabular}{@{}c@{}}\textbf{\hermesthree} \\ 8B\end{tabular} & \begin{tabular}{@{}c@{}}\textbf{\llamathreeone} \\ Instruct 8B\end{tabular} \\
    \hline
    \begin{tabular}{@{}c@{}}\textbf{AGIEval} \\ \textit{0-shot}\end{tabular} & \textbf{61.84} & 58.60 & \textbf{56.18} & 48.26 & \textbf{41.26} & 40.49 \\
    \begin{tabular}{@{}c@{}}\textbf{ARC-C} \\ \textit{0-shot}\end{tabular} & \textbf{69.45} & 66.04 & \textbf{65.53} & 63.40 & \textbf{58.11} & 55.12 \\
    \begin{tabular}{@{}c@{}}\textbf{ARC-E} \\ \textit{0-shot}\end{tabular} & \textbf{86.24} & 85.40 & 82.95 & \textbf{83.67} & \textbf{80.05} & 79.71 \\
    \begin{tabular}{@{}c@{}}\textbf{BoolQ} \\ \textit{0-shot}\end{tabular} & 88.93 & \textbf{89.52} & \textbf{88.04} & 87.76 & \textbf{84.95} & 84.01 \\
    \begin{tabular}{@{}c@{}}\textbf{BBH} \\ \textit{3-shot}\end{tabular} & 75.37 & \textbf{76.25} & 67.82 & \textbf{69.24} & \textbf{52.94} & 48.83 \\
    \begin{tabular}{@{}c@{}}\textbf{GPQA} \\ \textit{0-shot}\end{tabular} & \textbf{44.84} & 42.66 & 37.67  & \textbf{40.09} & 29.36 & \textbf{30.62} \\
    \begin{tabular}{@{}c@{}}\textbf{Hellaswag} \\ \textit{10-shot}\end{tabular} & \textbf{90.19} & 88.34 & \textbf{88.19} & 86.42 & \textbf{82.83} & 80.01 \\
    \begin{tabular}{@{}c@{}}\textbf{IFEval} \\ \textit{Strict}\end{tabular} & 84.87 & \textbf{87.09} & 81.21 & \textbf{87.25} & 62.25 & \textbf{80.15} \\
    \begin{tabular}{@{}c@{}}\textbf{MATH Lvl 5} \\ \textit{4-shot}\end{tabular} & 30.85 & \textbf{35.98} & 20.80 & \textbf{29.24} & ~7.48 & \textbf{~8.91} \\
    \begin{tabular}{@{}c@{}}\textbf{MMLU} \\ \textit{5-shot}\end{tabular} & 85.02 & \textbf{86.14} & 79.09 & \textbf{82.27} & 64.79 & \textbf{68.05} \\
    \begin{tabular}{@{}c@{}}\textbf{MMLU-PRO} \\ \textit{5-shot}\end{tabular} & 54.14 & \textbf{63.51} & 47.24 & \textbf{52.94} & 32.08 & \textbf{35.77} \\
    \begin{tabular}{@{}c@{}}\textbf{MT-Bench} \\ \textit{Avg.}\end{tabular} & ~8.93 & ~\textbf{9.17} & ~\textbf{8.99} & ~8.93 & ~8.27 & ~\textbf{8.39} \\
    \begin{tabular}{@{}c@{}}\textbf{MuSR} \\ \textit{0-shot}\end{tabular} & \textbf{48.26} & 47.58 & \textbf{50.67} & 47.08 & \textbf{43.52} & 38.23 \\
    \begin{tabular}{@{}c@{}}\textbf{OpenbookQA} \\ \textit{0-shot}\end{tabular} & \textbf{48.80} & 48.60 & \textbf{49.40} & 47.20 & \textbf{47.80} & 43.20 \\
    \begin{tabular}{@{}c@{}}\textbf{PiQA} \\ \textit{0-shot}\end{tabular} & \textbf{85.96} & 84.93 & \textbf{84.44} & 83.73 & 80.25 & \textbf{81.01} \\
    \begin{tabular}{@{}c@{}}\textbf{TruthfulQA} \\ \textit{MC2 0-shot}\end{tabular} & \textbf{65.57} & 64.83 & \textbf{63.29} & 59.91 & \textbf{58.69} & 53.99 \\
    \begin{tabular}{@{}c@{}}\textbf{Winogrande} \\ \textit{5-shot}\end{tabular} & 86.27 & \textbf{86.82} & 83.19 & \textbf{85.00} & 77.74 & \textbf{77.90} \\
    \end{tabular}
    \label{tab:performance}
\end{table}

\pagebreak

\bibliographystyle{abbrvnat}
\bibliography{references}

\appendix

\section{Generation Samples}

\begin{figure}[h!]
    \centering
    \begin{tcolorbox}
    \begin{tabular}{@{}p{0.125\textwidth}@{}p{0.875\textwidth}@{}}
    \textbf{System:} & \texttt{} \\[2ex]
    \textbf{User:} & \texttt{Testing, testing. Hello, who are you?} \\
    \end{tabular}
    \tcblower
    \begin{tabular}{@{}p{0.125\textwidth}@{}p{0.875\textwidth}@{}}
    \textbf{Assistant:} & \texttt{*looks around confused* I... I don't know who I am. Where am I? What's going on? I can't remember anything at all... *rubs head* My mind feels so foggy. What was the question again? I'm so disoriented...}
    \end{tabular}
    \end{tcolorbox}
    \caption{Empty system prompt generation (\hermesthree 405B)}
    \label{fig:gen:empty}
\end{figure}

\begin{figure}[h!]
    \begin{tcolorbox}
    \begin{tabular}{@{}p{0.125\textwidth}@{}p{0.875\textwidth}@{}}
    \textbf{System:} & \texttt{Please act as an impartial judge and evaluate the quality of the response provided by an AI assistant to the user question displayed below. Your evaluation should consider factors such as the helpfulness, relevance, accuracy, depth, creativity, and level of detail of the response. Your evaluation should focus on the assistant's answer to the second user question. Begin your evaluation by providing a short explanation. Be as objective as possible. After providing your explanation, you must rate the response on a scale of 1 to 10 by strictly following this format: "[[rating]]", for example: "Rating: [[5]]".} \\[2ex]
    \textbf{User:} & \texttt{<|The Start of Assistant A's Conversation with User|>} \\
    & \texttt{\#\#\# User: In the field of quantum physics, what is superposition, and how does it relate to the phenomenon of quantum entanglement?} \\
    & \\
    & ... \\
    & \\
    & \texttt{In summary, superposition is a fundamental property of quantum mechanics that allows for the existence of multiple states simultaneously, while entanglement is a phenomenon that arises from the superposition principle, where two or more particles become correlated in a way that is not possible in classical physics.} \\
    & \texttt{\#\#\# User: What assumptions have you made in your response? Are they valid?} \\
    & \\
    & ... \\
    & \\
    & \texttt{<|The End of Assistant A's Conversation with User|>}
    \end{tabular}
    \tcblower
    \begin{tabular}{@{}p{0.125\textwidth}@{}p{0.875\textwidth}@{}}
    \textbf{Assistant:} & \texttt{The assistant's response to the second question is thorough and well-structured. It identifies and evaluates several key assumptions made in the initial response, providing a nuanced perspective on the validity and limitations of these assumptions. The assistant acknowledges the widespread acceptance of quantum mechanics and the Copenhagen interpretation, while also recognizing the existence of alternative interpretations and ongoing debates in the field. It also addresses the applicability of quantum mechanics at different scales and the potential limitations of current mathematical formalism.} \\
    & \\
    & ... \\
    & \\
    & \texttt{Rating: [[8]]}
    \end{tabular}
    \end{tcolorbox}
    \caption{MT-Bench inspired reward modeling (\hermesthree 70B)}
    \label{fig:gen:judge}
\end{figure}

\begin{figure}[h!]
    \centering
    \begin{tcolorbox}
    \begin{tabular}{@{}p{0.125\textwidth}@{}p{0.875\textwidth}@{}}
    \textbf{System:} & \texttt{You are a conversational AI assistant that is provided a list of documents and a user query to answer based on information from the documents. You should always use grounded information in your responses, only answering from what you can cite in the documents. Cite all facts from the documents using <co: doc\_id></co> tags.}  \\ [2ex]
    \textbf{User:} & \texttt{What does Agent Instruct do?} \\
    & \\
    & \texttt{CONTEXT:} \\
    & \texttt{Document:0} \\
    & \texttt{Title: AgentInstruct: Toward Generative Teaching with Agentic Flows} \\
    & \texttt{Text: Synthetic data is becoming increasingly important for accelerating the development of language models ...} \\
    & \texttt{Document:1} \\
    & \texttt{Title: Advances in Emergency Medical Technology} \\
    & \texttt{Text: In recent years, emergency medical technology has seen significant advancements ...} \\
    & \\
    & \texttt{Document:2} \\
    & \texttt{Title: The Rise of Agentic Data Generation} \\
    & \texttt{Text: AgentInstruct is an agentic framework by Mitra et al. (2024), designed to generate large-scale ...}
    \end{tabular}
    \tcblower
    \begin{tabular}{@{}p{0.125\textwidth}@{}p{0.875\textwidth}@{}}
    \textbf{Assisstant:} & \texttt{AgentInstruct is an agentic framework designed to generate large-scale, diverse, and high-quality synthetic data. <co:2>The framework uses a sophisticated pipeline that transforms raw text into refined instructions through multiple stages of processing. The pipeline consists of four main steps: Seed Collection, Content Transformation, Seed Instruction Generation, and Instruction Refinement. Each stage in the pipeline is designed to enhance the complexity and quality of the generated instructions, ultimately creating a dataset that can be used to fine-tune language models.</co>} \\
    & \texttt{Cited Documents: 2} \\[2ex]
    \textbf{User:} & \texttt{What about model collapse?} \\[2ex]
    \textbf{Assistant:} & \texttt{Model collapse is a concern that has been raised by researchers in the context of using synthetic data for language model development. <co:0>Despite several successful use cases, researchers have pointed out that the use of synthetic data can lead to model collapse and drawbacks of imitating other models. This discrepancy in outcomes can be attributed to the fact that synthetic data varies in quality and diversity, and effective use of such data usually requires significant human effort in curating the data.</co>} \\
    & \texttt{Cited Documents: 0}
    \end{tabular}
    \end{tcolorbox}
    \caption{Retrieval augmented generation (\hermesthree 70B)}
    \label{fig:gen:rag}
\end{figure}

\begin{figure}[h!]
    \centering
    \begin{tcolorbox}
    \begin{tabular}{@{}p{0.125\textwidth}@{}p{0.875\textwidth}@{}}
    \textbf{System:} & \texttt{You should always respond in the following manner to all code problems:} \\
    & \\
    & \texttt{First, examine the problem by restating it, surrounded in the tags <RESTATEMENT> and </RESTATEMENT>. Next, the assistant will enclose all reasoning or insights within the tags <REASONING> and </REASONING> with each insight/reasoning surrounded in <THOUGHT\_N> </THOUGHT\_N> - where N is the index number of the thought. Then, after your reasoning and insights, form a step by step plan for the code you will produce, with <PLAN> </PLAN>, with each step being tagged <STEP\_N> </STEP\_N>, nested inside the <PLAN> tags, where N is the step number. After that, create Pydantic schema objects for any of the classes, functions, or other objects you may need or will use in your plan, surround those with <PYDANTIC\_SCHEMAS> </PYDANTIC\_SCHEMAS> and for each object schema, surround each with <SCHEMA\_N> </SCHEMA\_N> - indicating a separate distinct Pydantic object. Next, create a UML format diagram to explain the workflow of the code you will write, surrounded in <DIAGRAM> and </DIAGRAM> tags. Don't include direct calculations or code/functions in any of the mermaid diagram nodes. Always surround all text you place for the nodes with double quotes and if you need to use quotes inside an element, use single quotes instead, to make sure it doesn't get messy or error. Be detailed but only explain formulas, calculations, or code in natural language in your diagram nodes to ensure they compile successfully. Finally, inside the tags <REFLECTION> </REFLECTION> display an internal monologue, reflecting on the restatement, reasoning, plan, and diagram you've made. Critique it to ensure you have no blindspots for solving the problem, and if you do find it is wrong, note how you will edit your solution.} \\
    & \\
    & \texttt{Then the assistant can close the SCRATCHPAD tag, and proceed to provide a solution the problem, which should be done within <SOLUTION> </SOLUTION> tags.} \\
    &
    \texttt{After the solution, provide an explanation for all of the code, within <EXPLANATION> </EXPLANATION> XML tags.} \\
    & \\
    & \texttt{Finally, after the explanation, provide unit test code that is robust inside <UNIT\_TEST> </UNIT\_TEST> in the same programming language used to write the solution code.} \\[2ex]
    \textbf{User:} & \texttt{Write code to make a Discord bot that integrates the most popular LLM from huggingface as a chatbot in your server.}
    \end{tabular}
    \begin{tabular}{@{}p{0.125\textwidth}@{}p{0.875\textwidth}@{}}
    \textbf{Assistant:} & \texttt{<SCRATCHPAD>} \\
    & \texttt{<RESTATEMENT>} \\
    & \texttt{The task is to write Python code that will:} \\
    & \texttt{1. Automatically find the most popular LLM on Huggingface} \\
    & \texttt{2. Use that model as the basis for a chatbot} \\
    & \texttt{3. Create a Discord bot that uses this chatbot} \\
    & \texttt{</RESTATEMENT>} \\
    & \texttt{} \\
    & \texttt{<REASONING>} \\
    & \texttt{<THOUGHT\_1>} \\
    & \texttt{To find the most popular LLM on Huggingface, we can use the Huggingface API to retrieve the list of models and their respective download counts or stars. We can then sort the models based on these metrics to determine the most popular one.} \\
    & \texttt{</THOUGHT\_1>} \\
    & \\
    & ... \\
    & \\
    \end{tabular}
    \end{tcolorbox}
    \caption{Agentic coding (\hermesthree 70B)}
    \label{fig:gen:agentic_coding}
\end{figure}

\end{document}